\documentclass[runningheads]{llncs}
\usepackage{amsmath}
\usepackage{esvect}
\usepackage{bm}

\usepackage{float}
\usepackage[T1]{fontenc}
\usepackage{caption}
\usepackage{graphicx}
\usepackage{chngcntr}
\usepackage{subcaption}
\usepackage{multirow}
\counterwithout{figure}{section}   
\renewcommand{\thefigure}{\arabic{figure}}
\usepackage{float} 

\usepackage{algorithm}

\usepackage{amssymb}

\usepackage{colortbl}
\usepackage[colorlinks,urlcolor=blue,linkcolor=blue,citecolor=blue]{hyperref}
\usepackage{xcolor}
\usepackage{booktabs}

\usepackage{amsmath}
\usepackage{tgpagella}
\usepackage{todonotes}
\usepackage{algorithm} 
\usepackage{algpseudocode} 
\usepackage{amsmath}
\usepackage{amssymb}
\usepackage{graphicx,subcaption,adjustbox}
\usepackage{algorithm}
\usepackage{algorithmicx}
\usepackage{algpseudocode}
\usepackage{afterpage}
\usepackage{subcaption}
\usepackage{subcaption}
\usepackage{caption}

\algrenewcommand\algorithmicrequire{\textbf{Inputs:}}
\algrenewcommand\algorithmicensure{\textbf{Outputs:}}
\begin{document}

\title{I Detect What I Don’t Know: \\Incremental Anomaly Learning with Stochastic Weight Averaging-Gaussian for Oracle-Free Medical Imaging}

\author{
    Nand Kumar Yadav\inst{1} \and
    Rodrigue Rizk\inst{1} \and
    William C.~W.~Chen\inst{2} \and
    KC Santosh\inst{1}
}

\institute{AI Research Lab, Department of Computer Science \\
    \and
    \inst{}Biomedical \& Translational Sciences, Sanford School of Medicine \\
    University of South Dakota, Vermillion, SD 57069, USA \\
    \email{\{nand.yadav, rodrigue.rizk, william.chen, kc.santosh\}@usd.edu}
}

\maketitle

\begin{abstract}

Unknown anomaly detection in medical imaging remains a fundamental challenge due to the scarcity of labeled anomalies and the high cost of expert supervision. We introduce an unsupervised, oracle-free framework that incrementally expands a trusted set of normal samples without any anomaly labels. Starting from a small, verified seed of normal images, our method alternates between lightweight adapter updates and uncertainty-gated sample admission. A frozen pretrained vision backbone is augmented with tiny convolutional adapters, ensuring rapid domain adaptation with negligible computational overhead. Extracted embeddings are stored in a compact coreset enabling efficient $k$-nearest neighbor ($k$-NN) anomaly scoring. Safety during incremental expansion is enforced by dual probabilistic gates: a sample is admitted into the normal memory only if (i) its distance to the existing coreset lies within a calibrated z-score threshold, and (ii) its SWAG-based epistemic uncertainty remains below a seed-calibrated bound. This mechanism prevents drift and false inclusions without relying on generative reconstruction or replay buffers. 
Empirically, our system steadily refines the notion of “normality” as unlabeled data arrive, producing substantial gains over baselines. On COVID-CXR, ROC-AUC improves from $0.9489 \rightarrow 0.9982$ (F1: $0.8048 \rightarrow 0.9746$); on Pneumonia-CXR, ROC-AUC rises from $0.6834 \rightarrow 0.8968$; and on Brain MRI ND-5, ROC-AUC increases from $0.6041 \rightarrow 0.7269$ and PR-AUC from $0.7539 \rightarrow 0.8211$. These results highlight the effectiveness and efficiency of the proposed framework for real-world, label-scarce medical imaging applications~\footnote[1]{Source code is publicly available at: \url{https://github.com/USD-AI-ResearchLab/Incremental-learning-Anomaly-Detection}}.

\end{abstract}
%



\section{Introduction}
\label{sec:intro}
Anomaly detection, the task of identifying data points that deviate significantly from the normal data distribution, is a fundamental problem across a range of applications, including fraud detection, medical diagnostics, and cybersecurity. The ability to detect anomalies is critical, as in medical imaging, particularly in early diagnosis tasks like COVID-19 screening or tumor identification. Traditionally, anomaly detection methods rely on labeled anomalous data, which allows a model to learn to differentiate between normal and anomalous instances. However, obtaining labeled anomalous data is often difficult, expensive, and sometimes impossible, particularly in domains where anomalies are rare or novel. This limitation motivates the need for unsupervised anomaly detection methods, which do not require anomalous samples for training.

Before deep learning models, anomaly detection relied largely on methods such as One-Class SVMs \cite{scholkopf2001estimating} and $k$-NN~\cite{ramaswamy2000efficient}. One-Class SVM, for instance, learns a decision boundary around the normal data, classifying anything outside this boundary as an anomaly. Similarly, $k$-NN identifies anomalies by measuring the distance between a data point and its nearest neighbors, flagging points as anomalous that are far from their neighbors. Classical methods work on low-dimensional data, but struggle with high-dimensional data (e.g., images) where distance metrics break down due to the curse of dimensionality. These methods are highly sensitive to hyperparameters (e.g., SVM kernels, the choice of $k$ in $k$-NN) and often fail to capture complex relationships in image data, making them a poor fit for modern high-dimensional images.

To address these limitations, deep learning-based methods have gained traction in the field of anomaly detection. Autoencoders (AEs) and GANs~\cite{schlegl2019f} are now popular for anomaly detection in complex high-dimensional data. AEs learn a compact representation and try to reconstruct the input; samples with a large reconstruction error are treated as anomalies because they do not look “normal” to the model. Variants like VAEs~\cite{kingma2014auto} and denoising AEs~\cite{vincent2008extracting} boost robustness and generalization, especially with noisy inputs. However, while AEs have shown success, they suffer from overfitting to the normal data distribution. This overfitting problem limits their practical use in detecting novel or rare anomalies. Furthermore, the performance of AEs heavily depends on the quality and quantity of normal data, and they may struggle to detect anomalies that deviate substantially from typical patterns in the normal dataset.

GANs~\cite{goodfellow2014generative} and AEs have been widely applied to anomaly detection. These models are typically trained on normal data with the goal of reconstructing or generating samples that resemble the normal distribution, such as AnoGAN~\cite{schlegl2019f}. Anomalies are then detected by identifying instances that the model reconstructs or generates poorly. However, such approaches suffer from several drawbacks. AEs often learn to reconstruct anomalies as well as normal data, leading to reduced discriminative power, while GANs are prone to training instability, sensitivity to hyperparameters, and require large amounts of normal data for effective training. These limitations hinder their scalability and robustness in real-world anomaly detection tasks.

More recently, representation-based methods such as PatchCore \cite{roth2022towards} have demonstrated superior performance by leveraging pretrained networks to extract patch-level embeddings of normal data. Instead of relying on generative reconstruction, PatchCore builds a memory bank of patch embeddings and detects anomalies by measuring the distances between test patches and their nearest neighbors in this memory. This avoids the pitfalls of generative approaches and enables robust, high-resolution anomaly localization.

Alongside these advances, incremental learning (IL) has emerged as a complementary strategy to reduce the reliance on large amounts of labeled anomalous data. By iteratively selecting the most informative or uncertain instances for labeling, IL improves model performance with minimal annotation effort. However, applying IL to anomaly detection remains challenging, as anomalies are rare or absent in training, making uncertainty estimation non-trivial.

To address these limitations, we propose a novel oracle-free, unsupervised anomaly detection framework that learns from a small initial set of normal samples and iteratively updates its normality model over time, entirely without requiring anomalous labels.

We build upon the PatchCore ~\cite{roth2022patchcore} architecture, known for its efficient, non-generative coreset-based detection, and integrate it with: (a) incremental learning to update the memory bank with novel, high-confidence normal samples, (b) Stochastic Weight Averaging-Gaussian (SWAG) ~\cite{maddox2019simple} to estimate epistemic uncertainty for gating sample inclusion. Our contributions are as follows:
\begin{itemize}
    \item A novel IL-PatchCore+SWAG framework for unsupervised anomaly detection that incrementally expands the normal class distribution using adapter-based feature refinement and SWAG-driven uncertainty gating.
    \item A safe sample selection strategy combining distance-to-core and epistemic variance thresholds to prevent contamination of the normal memory bank.
    \item Demonstration of high detection performance on real-world datasets: COVID-CXR, Pneumonia Chest X-ray, and Brain MRI, outperforming strong baselines.
 
\end{itemize}

\section{Related Works}
\subsection{Classical Anomaly Detection Methods}
Classical anomaly detection techniques such as One-Class SVM \cite{scholkopf2001estimating} and $k$-NN~\cite{ramaswamy2000efficient} were among the earliest unsupervised approaches: the former encloses normal data within a decision boundary and flags points outside it, while the latter measures a sample’s distance to its nearest neighbors and treats large distances as anomalous. Despite their historical impact, these methods face well-known limitations. In high-dimensional settings, distance metrics become less discriminative, eroding the effectiveness of $k$-NN style detectors; One-Class SVM is notably sensitive to kernel and regularization choices, often demanding costly cross-validation; and both families struggle to model the complex, structured dependencies present in images or sequences. Consequently, although they remain practical for simple, low-dimensional problems, their performance degrades on modern, high-dimensional data, motivating the shift toward deep learning–based methods that can better capture rich feature hierarchies.

\subsection{Deep Learning-Based Anomaly Detection}
Deep learning approaches, particularly AEs and GANs, have become widely adopted for anomaly detection in high-dimensional and complex domains such as image data.

\bigskip
\noindent \textbf{AEs.} AEs are neural networks that learn compressed latent representations of input data with the objective of reconstructing the original inputs. Anomaly scores are typically derived from reconstruction error, with higher errors indicating a greater likelihood of anomalousness. Variants such as VAEs \cite{kingma2014auto} and Denoising AEs \cite{vincent2008extracting} have been proposed to improve reconstruction fidelity and generalization. Despite their popularity, AEs exhibit notable limitations. They are prone to reconstruction bias, often overfitting to normal data and thereby reducing their ability to generalize to unseen anomalies \cite{an2015variational}. Furthermore, their effectiveness is strongly dependent on the availability of abundant, high-quality normal data, which limits their applicability in settings with scarce or imbalanced datasets. 

\bigskip
\noindent \textbf{Generative Adversarial Networks (GANs) and PatchCore.}
GANs, composed of a generator and discriminator pair, have also been investigated for anomaly detection. The generator aims to approximate the distribution of normal data, while the discriminator distinguishes between real and synthetic samples, with anomalies identified when they fall outside the modeled distribution~\cite{scholkopf2001estimating}. Despite their potential, GAN-based approaches suffer from several limitations. They are prone to training instability and mode collapse, which can result in incomplete coverage of the normal distribution \cite{goodfellow2014generative}. Moreover, training an effective GAN typically requires large volumes of normal data, which is often impractical in real-world anomaly detection scenarios. GANs are highly sensitive to hyperparameters, making them computationally expensive and difficult to optimize reliably. These drawbacks restrict their scalability and limit their robustness as a practical solution for anomaly detection. PatchCore \cite{roth2022patchcore} is a state-of-the-art approach for unsupervised anomaly detection that avoids the need for generative modeling. Instead of reconstructing or synthesizing images (as in AEs or GANs), PatchCore directly leverages deep features from pretrained convolutional networks. It extracts multi-scale patch-level features from training images and stores a compact subset in a memory bank using coreset subsampling. This design has several advantages: it avoids the training instability of GANs, over-generalization issues of AEs, and requires only normal data for building the memory bank. 

Consequently, PatchCore achieves high detection accuracy with efficient inference, making it well-suited for industrial and medical anomaly detection tasks. However, PatchCore also suffers from large memory requirements during the feature storage and learning phase, since even the coreset can grow significantly with data size. To address this drawback, we propose incorporating IL to select only the most informative samples from the pool. Additionally, we use SWAG-based uncertainty estimation to guide the selection of the most appropriate samples. This makes the method more computationally efficient, memory-friendly, and better suited for real-time deployment, while maintaining high detection performance.

\noindent \textbf{IL in Anomaly Detection.} 
Incremental (continual) learning updates a model as data arrive over time while avoiding catastrophic forgetting. In anomaly detection this is especially challenging: (i) the data distribution drifts (both covariate and semantic), so a static threshold or memory quickly becomes obsolete; (ii) anomalies are rare and diverse, so incremental updates risk \emph{contaminating} the normal model when labels are absent; (iii) strict memory and latency budgets limit rehearsal of past data, exacerbating the stability–plasticity trade-off; and (iv) most continual-learning methods assume task or class labels and class-balanced streams, assumptions that rarely hold in one-class settings. Consequently, effective incremental anomaly detection requires drift-aware calibration, bounded exemplar memories (e.g., coreset selection), to preserve prior normality, ideally under weak or no anomaly labels \cite{parisi2019continual,rebuffi2017icarl,li2016learning,delange2021continual}. This perspective aligns with our design: we update the detector online using a frozen normal memory with cautious admission criteria and lightweight adapter tuning, thereby maintaining past competence while adapting to new normal regimes without relying on labeled anomalies. 

\bigskip
\noindent \textbf{Anomaly Detection Without Anomalous Training Data.}
Recent methods remove the need for anomalous training labels by first modeling normality and then flagging deviations. Self-supervised one-class objectives and patch-based methods fall into this camp: they learn feature spaces where normal samples are compact and anomalies appear distant. Despite promising results, purely unsupervised pipelines remain vulnerable to pool contamination (i.e., inadvertent inclusion of anomalies), and they rarely provide a principled way to \emph{expand} the normal manifold once deployment uncovers new, legitimate variability. These limitations motivate our approach. Starting from a small, trusted seed of normals, we enlarge the normal set incrementally. At each step, a candidate is admitted only when its PatchCore distance is statistically consistent with the seed distribution and its epistemic uncertainty estimated via SWAG is low. This dual evidence gate improves robustness to contamination and offers a principled path for safe, label-free growth of the memory bank, addressing both data efficiency and long-term stability of the detector.

\section{Motivation} 
This work is motivated by the following question: \emph{can we safely grow the set of normals without ever consulting an oracle?} Our answer is to use pool-based selection with IL under a \emph{dual-evidence gate}. Concretely, we start from a small, trusted seed of normals, build a compact PatchCore memory via coreset subsampling \cite{roth2022towards}, and then expand that memory only with pool samples that simultaneously (i) score low under PatchCore  $k$-NN distance after seed-based z-normalization (statistical evidence), and (ii) exhibit low epistemic uncertainty under a SWAG posterior over the adapter \cite{maddox2019simple} (Bayesian evidence). In contrast to traditional active learning~\cite{bouguelia2018agreeing}, where high uncertainty triggers an expensive query, uncertainty here acts as a \emph{filter}: items are admitted \emph{only if} the model is confident they lie inside the current normal manifold. The adapter is then updated briefly on the admitted samples. This incremental active~\cite{santosh2020ai},\cite{santosh2023active} recipe is attractive for three reasons. First, it is \emph{oracle-free and data-efficient}: a tiny seed bootstraps a detector that improves as new, confident normals arrive. Second, it is \emph{robust}: the dual gate limits contamination from hidden anomalies, while seed-relative z-scores and periodic recalibration provide drift awareness. Third, it is \emph{scalable}: PatchCore avoids generative training. and under mild separation/concentration assumptions, adding verified normals can only lower expected normal scores, so in the clean limit the normal manifold sharpens, giving a principled path to stable, label-free improvement over time.

In short, our motivation is practical and structural: replace an impractical human oracle with a Bayesian filter, couple it with a conservative statistical test relative to a trusted seed, and grow normality incrementally with bounded risk so anomaly detectors remain accurate, robust, and maintainable under real deployment constraints.

\section{Proposed Method}
\begin{figure}[tbp]
  \centering
  \includegraphics[width=0.95\linewidth]{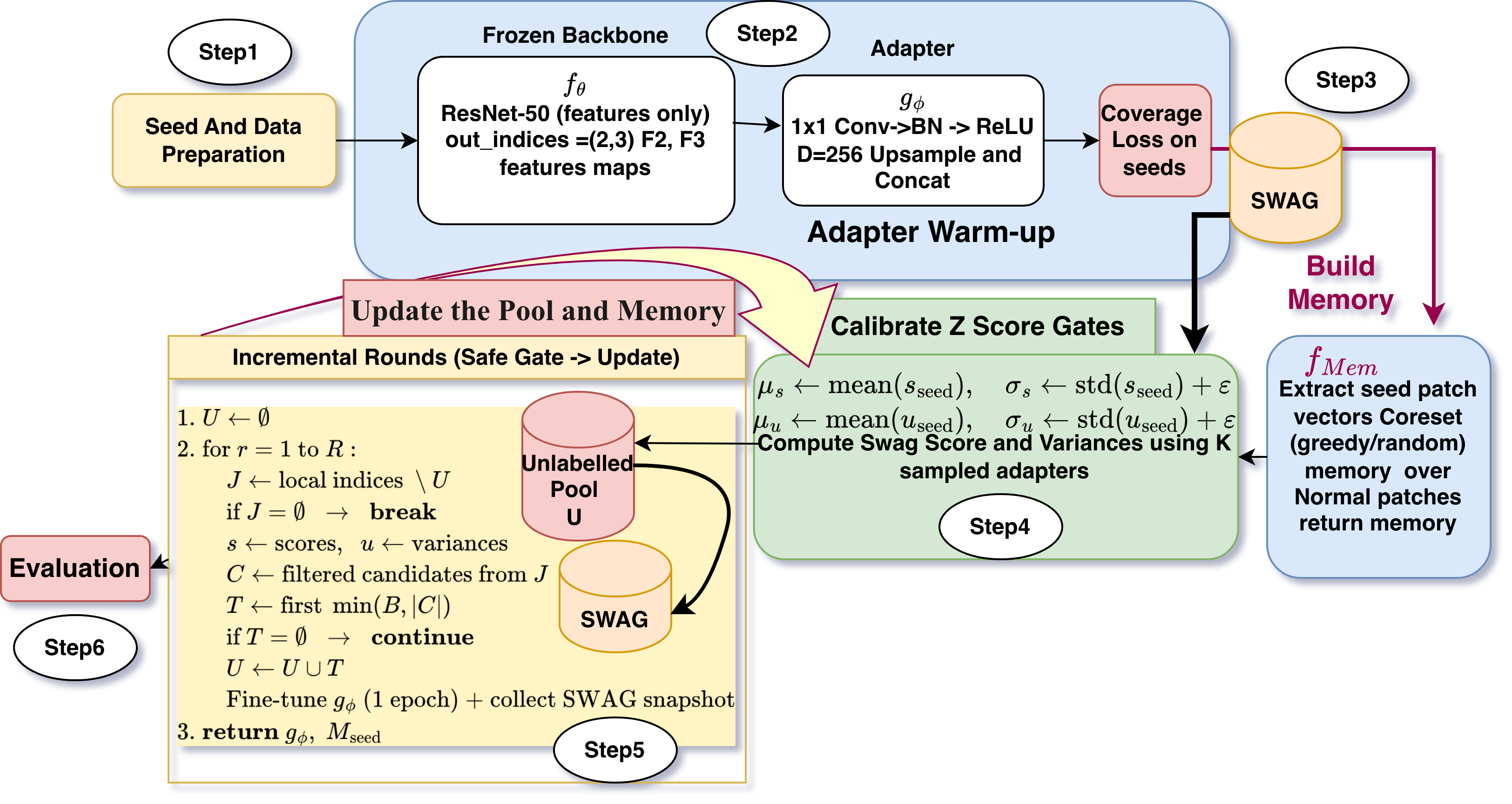}
  \caption{Pipeline of the proposed anomaly-detection framework.}
  \label{fig:pro}
\end{figure}

We consider image anomaly detection where training images come primarily from a $\emph{normal}$ class, while the test set contains both normal and anomalous samples. 
We adopt a PatchCore\mbox{-}style non\mbox{-}parametric detector built on a frozen, pretrained ResNet backbone $f_\theta$. Given an image $x\in[0,1]^{C\times H\times W}$, multi\mbox{-}scale feature maps are extracted and passed through a lightweight adapter $g_\phi$ that projects each scale to a common width; the outputs are bilinearly aligned, with channel\mbox{-}concatenated, and $\ell_2$-normalized to yield per\mbox{-}location patch embeddings. A memory bank $\mathcal{M}$ of normal embeddings is constructed from trusted seed data via farthest\mbox{-}first (coreset) selection.
 We incorporate SWAG-based uncertainty estimates to complement distance scoring. From an unlabeled pool, additional pseudo-normal samples are admitted only when statistical and epistemic evidence supports their inclusion, enabling a cautious but steady enlargement of the normal manifold. This setting highlights the fundamental difficulty of anomaly detection: while normal data are abundant and well-characterized, anomalies remain rare, diverse, and context dependent, making them unsuitable for a supervised approach. On the other hand, unsupervised learning from large data pools~\cite{11050525} risks contamination, since even a small fraction of anomalies can distort the learned statistics. Our formulation begins with a small, trusted seed of normal data and carefully grows outward, admitting new samples under principled uncertainty checks. This progressive, evidence-driven expansion of normality ensures robustness, efficiency, and stability, and sets the stage for our algorithms (Algos.~\ref{alg:seed_data}–\ref{alg:al_eval}) that formalize each component of this process. The overall pipeline is depicted in Fig.~\ref{fig:pro}.

Our method operates over iterative rounds. Each round: (1) scores incoming unlabeled data using PatchCore-based $k$-NN anomaly scoring, (2) uses SWAG to estimate uncertainty on these samples, (3) selects high-confidence normal candidates based on: distance thresholding to the coreset, and epistemic uncertainty z-score relative to seed distribution, (4) admits selected samples into the coreset memory, and (5) updates the adapter modules (only) to improve representations.
The process begins with a seed set of trusted normal samples as illustrated in Algo.~\ref{alg:seed_data}. Furthermore, we use an adapter for learning the normal samples, as shown in Algorithm~\ref {alg:warmup}. Simultaneously, we populate a prototype memory by coreset selection from the seed embeddings as illustrated in Algo.~\ref{alg:memory}. This memory serves as a compact summary of the normal distribution, a reference structure against which all future candidates can be evaluated. Crucially, in our implementation, this seed memory remains fixed for the entirety of training, providing a stable foundation that prevents contamination by the pool. Even if a sample looks close to what we’ve seen before, the model also has to admit what it doesn’t know. A point might sit near our “normal” prototypes, but if the model is highly uncertain about its representation, admitting it could be risky. To address this, we equip the adapter with a Bayesian posterior approximation using SWAG, which maintains a distribution over its weights. By sampling multiple parameter realizations and scoring each candidate, we estimate both the mean distance $\mu(x)$ and variance $v(x)$ as illustrated in Algo.~\ref{alg:calib}. The mean captures how closely a sample resembles normal patterns, while the variance captures epistemic uncertainty: how consistently the model agrees on this assessment.
\begin{algorithm}[tbp]
\caption{(Step 1): Seed \& Data Preparation}
 \label{alg:seed_data}
\begin{algorithmic}[1]
\Require Normal dir $\texttt{dir}_0$, Anomaly dir $\texttt{dir}_1$, image size $S$, color $\in \{\texttt{L},\texttt{RGB}\}$
\State $\mathcal{X}\leftarrow$ load all images from $\texttt{dir}_0,\texttt{dir}_1$; resize to $(S,S)$; scale to $[0,1]$
\State Label mapping: $\texttt{NORMAL}\mapsto 0$, $\texttt{ANOMALY}\mapsto 1$
\State $\mathcal{I}_0 \leftarrow \{i: y_i=0\}$ \Comment{all normals}
\State Shuffle $\mathcal{I}_0$; \quad $n \leftarrow \max\!\big(1, \lfloor 0.30\cdot|\mathcal{I}_0|\rceil\big)$
\State $\mathcal{I}_{\text{seed}} \leftarrow$ first $n$ of $\mathcal{I}_0$;\quad $\mathcal{I}_{\text{rest}} \leftarrow \mathcal{I}_0\setminus \mathcal{I}_{\text{seed}}$
\State $\mathcal{I}_{\text{anom}} \leftarrow \{i: y_i=1\}$;\quad $\mathcal{I}_{\text{pool}} \leftarrow \mathcal{I}_{\text{rest}} \cup \mathcal{I}_{\text{anom}}$
\State \Return $(\mathcal{X}, \mathcal{I}_{\text{seed}}, \mathcal{I}_{\text{pool}})$
\end{algorithmic}
\end{algorithm}

The central decision is then governed by a \emph{z-score gate}. For each candidate, we normalize its score and uncertainty relative to the seed distribution:
\begin{equation}
    z_s(x) = \frac{\mu(x) - \mu_s}{\sigma_s}, \qquad
    z_u(x) = \frac{v(x) - \mu_u}{\sigma_u}.
\end{equation}
The sample is qualified as a pseudo-normal only when $z_s(x)$ and $z_u(x)$ are at most $1.0$. If no candidates pass, the gates may be relaxed once to $1.5\sigma$. Unlike earlier designs, there is no borderline band or fusion ranking: acceptance is a strict intersection of score and uncertainty gates. This mechanism enforces selective trust, ensuring that only statistically safe samples are admitted.
Accepted samples are then used for a brief adapter update, typically a single compact epoch, and the new weights are snapshotted into the SWAG ensemble. This incremental tuning allows the adapter to expand its coverage of $P_N$ without ever altering the frozen seed memory. The process repeats for several rounds until the pool is exhausted or no new candidates are admitted. In this way, the system grows cautiously: expanding its representation of normality while keeping the calibration anchored to the uncontaminated seed, the procedure is illustrated in Algo.~\ref{alg:active_rounds}.

Finally, evaluation is straightforward as illustrated in Algo.~\ref{alg:al_eval}. A test image is scored by PatchCore distances to the seed memory, using the most up-to-date adapter and SWAG ensemble. Anomalies emerge as those samples whose scores or uncertainties exceed the calibrated bounds. Each component plays a precise role: the seed set prevents initial contamination; the adapter tailors the latent space to the domain; SWAG quantifies epistemic uncertainty; the dual z-score gates prevent unsafe admissions; and the iterative loop allows gradual, self-supervised growth. What emerges is a robust, oracle-free learner: a model that begins with modest seeds, acknowledges its own uncertainty, and safely expands its knowledge of normality through IL.

\paragraph{Implications.}
Let 
\[
\beta \;=\; 
\Pr_{x\sim P_N}\!\big(z_s(x)\le\tau_z \,\wedge\, z_u(x)\le\tau_z\big)
\]
denote the acceptance probability of the dual gate for \emph{true normal} samples, 
with an optional one-time relaxation $\tau_z\!:1.0\!\rightarrow\!1.5$ if required. 
In the {strict\_normal\_only} =\textbf{True}, configuration 
(i.e., oracle-assisted mode), only gate-passing items whose ground-truth 
label is $y{=}0$ are used to update the adapter $g_\phi$ and to expand the 
PatchCore memory~$\mathcal{M}$. 
Each admitted batch therefore contributes $\Theta(\beta)$ genuine normal 
samples, incurs \emph{zero} anomaly contamination, and follows the standard 
finite-sample error decay $O(1/\!\sqrt{m})$ as the memory size $m$ increases. 
The dual gate serves as a conservative prefilter, requiring both low normalized 
distance ($z_s$) and low uncertainty ($z_u$), thereby improving sample 
efficiency while maintaining safety. 
Uncertainty is ignored only in degenerate cases where $\sigma_u$ is 
numerically negligible.

\begin{algorithm}[tbp]
\caption{(Step 2): Initialize Model and Warm-Up Adapter}
\label{alg:warmup}
\begin{algorithmic}[1]
\Require Frozen backbone $f_\theta$, adapter $g_\phi$ (per-scale $1{\times}1$ Conv+BN+ReLU, dim $D{=}256$), seed indices $\mathcal{I}_{\text{seed}}$, epochs $E$, lr $\eta$, prototype budget $K$
\Ensure Warmed $g_\phi$ and initialized SWAG
\State Set $f_\theta$ to \textbf{eval}; initialize $g_\phi$
\State Build seed vectors $V_{\text{seed}}$ \textbf{(no per-image cap, native grid)}:
\Statex \hspace{0.8em}for $x\in\mathcal{I}_{\text{seed}}$: get multi-scale $\{F_\ell\}$; project $A_\ell{=}g_\phi^\ell(F_\ell)$;
upsample each $A_\ell$ to the \textbf{common native size} $(H_{\max},W_{\max})$ of that image; 
concatenate channel-wise to $A$; flatten and $\ell_2$-normalize to get $Q$; append all $Q$ to $V_{\text{seed}}$.
\State $\rho \leftarrow \min(1, K/|V_{\text{seed}}|)$; \quad $P \leftarrow \textsc{CoresetGreedy}(V_{\text{seed}}, \rho)$
\For{$e=1..E$} \Comment{backbone frozen}
  \For{minibatches of $x\in\mathcal{I}_{\text{seed}}$}
    \State Extract $Q$ as above; \ \ $\mathcal{L}\leftarrow \frac{1}{|Q|}\sum_i \min_{\mathbf{p}\in P}\|\mathbf{q}_i-\mathbf{p}\|_2$
    \State Update $g_\phi$ with Adam$(\text{lr}{=}\eta)$ on $\mathcal{L}$
  \EndFor
\EndFor
\State Initialize SWAG on $g_\phi$ and take \textbf{two} snapshots
\State \Return warmed $g_\phi$, SWAG state
\end{algorithmic}
\end{algorithm}

 \begin{algorithm}[tbp]
\caption{(Step 3): Build Initial Normal Memory (PatchCore)}
\label{alg:memory}
\begin{algorithmic}[1]
\Require $f_\theta$, warmed $g_\phi$, seed set $\mathcal{I}_{\text{seed}}$, coreset ratio $\rho_m$, grid $16{\times}16$, cap $\tau$
\Ensure Initial memory $\mathcal{M}_0$
\State Extract $V_{\text{seed}}$ as in Alg.~\ref{alg:warmup}; \quad $\mathcal{M}_0 \leftarrow \textsc{CoresetGreedy}(V_{\text{seed}}, \rho_m)$
\State \Return $\mathcal{M}_0$ \Comment{Note: in incremental rounds we \emph{grow/rebuild} memory with accepted normals}
\end{algorithmic}
\end{algorithm}

\begin{algorithm}[tbp]
\setcounter{algorithm}{3}
\caption{(Step 4): Calibrate Z-Score Gates (on seeds)}
\label{alg:calib}
\begin{algorithmic}[1]
\Require $f_\theta$, $g_\phi$, memory $\mathcal{M}_0$, seed set $\mathcal{I}_{\text{seed}}$, $k$-NN $k{=}3$, top-$q$ (e.g.\ $0.03\!-\!0.05$), SWAG samples $K$
\Ensure $(\mu_s,\sigma_s,\mu_u,\sigma_u)$ and flag $\texttt{useU}$
\State For each $x\in\mathcal{I}_{\text{seed}}$: compute $s(x)$ via PatchCore top-$q$ rule with $\mathcal{M}_0$
\State For each $x\in\mathcal{I}_{\text{seed}}$: sample $K$ SWAG adapters, get $u(x){=}\mathrm{Var}[s^{(k)}(x)]$
\State $\mu_s,\sigma_s \leftarrow \mathrm{mean/std}\big(\{s(x)\}\big)$; \quad $\mu_u,\sigma_u \leftarrow \mathrm{mean/std}\big(\{u(x)\}\big)$
\State $\texttt{useU} \leftarrow (\sigma_u > 10^{-6})$ \Comment{if SWAG variance is numerically non-zero}
\State \Return $(\mu_s,\sigma_s,\mu_u,\sigma_u,\texttt{useU})$
\end{algorithmic}
\end{algorithm}

\begin{algorithm}[tbp]
\caption{(Step 5): Incremental Rounds (Growing Memory + Gated Ranking + Checkpointing)}
\label{alg:active_rounds}
\begin{algorithmic}[1]
\Require Backbone $f_\theta$ (frozen), adapter $g_\phi$, SWAG state; pool $\mathcal{I}_{\text{pool}}$; rounds $R$; budget $B$;
calibration $(\mu_s,\sigma_s,\mu_u,\sigma_u,\texttt{useU})$; thresholds $\tau_z{=}1.0\!\rightarrow\!1.5$ (one relaxation);
ranking mode $\in\{\texttt{boundary},\texttt{uncert}\}$; strict-normal-only flag (\texttt{False}/\texttt{True}); resume policy $\in\{\texttt{best\_so\_far},\texttt{last}\}$.
\Ensure Final adapter $g_\phi^{\star}$ and memory $\mathcal{M}^{\star}$.

\State $\mathcal{U}\leftarrow\emptyset$ \Comment{used local pool indices}; \quad $\mathcal{A}\leftarrow\emptyset$ \Comment{accepted normal indices}
\State Save \texttt{best\_overall} checkpoint of $g_\phi$ using a metric (validation AUC if available, else $\overline{s}_{\text{pool}}{-}\overline{s}_{\text{seed}}$ on a fixed subset)

\For{$r=1$ \textbf{to} $R$}
  \State \textbf{Resume} $g_\phi$ from \texttt{best\_overall} (or \texttt{last}) per policy
  \State \textbf{Build memory} from seeds $\cup$ accepted normals:
        $\mathcal{M}_r \!\leftarrow\! \textsc{CoresetGreedy}(V(\mathcal{I}_{\text{seed}}\cup\mathcal{A}),\rho_m)$
  \State $\mathcal{J}\!\leftarrow\!\{0,\dots,|\mathcal{I}_{\text{pool}}|\!-\!1\}\setminus\mathcal{U}$; \quad 
         \If{$\mathcal{J}=\emptyset$} \textbf{break} \EndIf

  \State Score pool with $\mathcal{M}_r$: $\{s_j\}_{j\in\mathcal{J}}$; 
         if $\texttt{useU}$ then also $\{u_j\}_{j\in\mathcal{J}}$ else $u_j{\leftarrow}0$
  \State Normalize: $z^s_j{=}(s_j{-}\mu_s)/\sigma_s$; \quad $z^u_j{=}(u_j{-}\mu_u)/\sigma_u$
  \State Safe set:
         $\mathcal{C}{\leftarrow}\{j\in\mathcal{J}\mid z^s_j\!\le\!\tau_z \wedge (\neg\texttt{useU} \lor z^u_j\!\le\!\tau_z)\}$
  \If{$\mathcal{C}=\emptyset$} set $\tau_z{\leftarrow}1.5$ and recompute $\mathcal{C}$; 
       \If{$\mathcal{C}=\emptyset$} \textbf{continue} \EndIf \EndIf

  \State \textbf{Rank} in $\mathcal{C}$:
  \Statex \hspace{1em}\texttt{boundary}: sort by $s_j$ (desc); \quad \texttt{uncert}: sort by $z^u_j$ (desc)
  \State $\mathcal{T}{\leftarrow}$ top-$\min(B,|\mathcal{C}|)$ locals after ranking
  \State Mark used: $\mathcal{U}{\leftarrow}\mathcal{U}\cup\mathcal{T}$; 
        \quad map to globals $\mathcal{G}{\leftarrow}\{\mathcal{I}_{\text{pool}}[j]\mid j\in\mathcal{T}\}$

  \If{\textbf{strict-normal-only}}
     \State Filter true normals: $\mathcal{G}{\leftarrow}\{g\in\mathcal{G}\mid y_g{=}0\}$
     \If{$|\mathcal{G}|{=}0$} 
        \State Save \texttt{last}; \textbf{continue} 
     \EndIf
  \Else
     \State \textit{// Unsupervised mode (no labels used from the pool)}
     \State Treat all selected samples as presumed normals: $\mathcal{G}{\leftarrow}\mathcal{T}$
  \EndIf

  \State \textbf{Fine-tune} $g_\phi$ for 1 epoch on $\mathcal{G}$ with prototype loss (backbone frozen); take a SWAG snapshot
  \State \textbf{Grow memory}: $\mathcal{A}{\leftarrow}\mathcal{A}\cup\mathcal{G}$
  \State \textbf{Checkpoint}: save \texttt{last}; recompute metric; if improved, update \texttt{best\_overall}
\EndFor

\State Restore $g_\phi^{\star}$ from \texttt{best\_overall}; \quad build $\mathcal{M}^{\star}\!\leftarrow\!V(\mathcal{I}_{\text{seed}}\cup\mathcal{A})$
\State \Return $g_\phi^{\star},\,\mathcal{M}^{\star}$
\end{algorithmic}
\vspace{-0.4em}
\end{algorithm}


\begin{algorithm}[tbp]
\caption{(Step 6): Evaluate on Held-out Test}
\label{alg:al_eval}
\begin{algorithmic}[1]
\Require Test loader $\mathcal{D}_{\text{test}}$, $f_\theta, g_\phi, \mathcal{M}$, labels $\{y_i\}$
\State Compute image-level anomaly scores $\{s_i\}$ via PatchCore (top-$q$ of $k$-NN distances to $\mathcal{M}$)
\If{\,$\{y_i\}$ contains both classes}
  \State Build ROC; $\text{AUC}\gets \text{area under ROC}$
  \State $t^\star \gets \arg\max_t \big(\mathrm{TPR}(t)-\mathrm{FPR}(t)\big)$ \Comment{Youden’s $J$}
  \State $\hat{y}_i \gets \mathbb{1}[s_i \ge t^\star]$; report ACC, Precision, Recall, F1
  \State (Optional) save ROC/PR plots and example heatmaps
\Else
  \State Skip thresholded metrics (no positives/negatives)
\EndIf
\end{algorithmic}
\end{algorithm}

\begin{theorem}[Oracle-assisted zero-contamination under dual-gate prefilter]
\label{thm:oracle_assisted}
Assume an oracle-assisted variant of Alg.~\ref{alg:active_rounds}, 
implemented conceptually via {strict\_normal\_only} =\textbf{True}. 
After the dual gate $(z_s,z_u)$ selects candidates satisfying 
$z_s(x)\!\le\!\tau_z$ and $z_u(x)\!\le\!\tau_z$, 
the ground-truth label $y(x)$ is queried, and only samples with $y{=}0$ are 
admitted for adapter updates and memory expansion. 
Then the contamination rate of the memory bank,
\[
\alpha \;=\; \Pr\!\big[x\in\mathcal{M} \,\wedge\, y(x){=}1\big],
\]
is identically $\alpha = 0$ for all active-learning rounds, 
independent of the gate threshold relaxation and SWAG sampling settings.
\end{theorem}
\noindent\textbf{\textit{Proof sketch.}} 
By construction, the oracle-assisted update rule admits only samples with 
$y{=}0$. 
Hence $\Pr[x\in\mathcal{M}\land y(x){=}1]=0$ for all rounds. 
The dual gate $(z_s,z_u)$ merely determines which items are 
\emph{considered}; 
the oracle filter decides which are actually \emph{added}.

\begin{proposition}[Oracle-free dual-gate bound (image level)]
\label{prop:oracle_free}
If the oracle filter is disabled (\texttt{strict\_normal\_only}=\textbf{False}) 
and there exist separation margins $\gamma_s,\gamma_u>0$ such that for all 
anomalous $x\!\sim\!P_A$ either 
$z_s(x)\!\ge\!1+\gamma_s$ or $z_u(x)\!\ge\!1+\gamma_u$, 
then, with $K$ SWAG samples and unbiased, finite-variance estimators of 
$d(x)$ and $u(x)$, the probability that an anomaly passes the dual gate 
$z_s(x)\!\le\!1$ and $z_u(x)\!\le\!1$ satisfies
\[
\Pr_{x\sim P_A}\!\big[\text{admit}(x)\big]
\;\le\;
\delta 
\;+\;
\exp\!\left(
   -\,cK \cdot 
   \min\!\Big\{\tfrac{\gamma_s^2}{\sigma_s^2},
                \tfrac{\gamma_u^2}{\sigma_u^2}\Big\}
\right),
\]
for some universal constant $c>0$, where $\delta$ accounts for 
finite-sample concentration. 
In degenerate cases where $\sigma_u\!\approx\!0$, 
the decision reduces to the single-score gate on $z_s$ alone.
\end{proposition}

\paragraph{Incremental Learning with and without an Oracle.}
Classical IL forwards uncertain samples to an oracle~\cite{yin2024efficient}. The uncertainty acts \emph{as a filter}: only low-distance, low-uncertainty candidates are eligible. 
With the oracle filter (our default), memory growth is provably contamination-free (Thm.~\ref{thm:oracle_assisted}); without it, the dual gate provides the probabilistic safety in Prop.~\ref{prop:oracle_free}.

\section{Experimental Setup, Results, and Discussion}
The experimental setup uses a frozen ResNet-50 backbone with selected scales projected by an adapter using $1{\times}1$ Conv+BN+ReLU layers to a width of 256, normalized channel-wise before patch flattening. During warm-up, prototypes are formed on seed normals at native resolution with a prototype-coverage loss, and SWAG is initialized with adapter snapshots. For memory, compact $16{\times}16$ patch grids are used for images, while scoring uses full-resolution aligned maps. SWAG generates sampled adapters by perturbing floating weights, and dual gating with z-scores on PatchCore scores and variance filters candidates, with relaxed thresholds if necessary. Selection ranks safe items either by boundary score or uncertainty, then adapters are fine-tuned strictly on accepted normals with small updates, followed by new SWAG snapshots and memory rebuilds. A best-so-far checkpoint is maintained based on validation AUC. Scoring employs PatchCore with $k=3$ and top-3–5\% patch fraction, and performance is reported with ROC-AUC, PR-AUC, and thresholds like Youden’s $J$~\cite{youden1950index} or target precision.

\paragraph{Hyperparameter Settings.}
All experiments were implemented using random seeds (123--127) to ensure reproducibility. 
The learning rate was set to $1\times10^{-4}$ during initial adapter training and reduced to $3\times10^{-5}$ for incremental fine-tuning, 
providing stable convergence while preserving previously learned representations. 
We used a batch size of 32, a coreset memory ratio of 0.3, and performed five active learning rounds with a budget of 200 samples per round for the Chest-xray dataset and 500 for the Brain MRI dataset. The SWAG uncertainty estimation was configured with a noise scale of 0.02 and $K{=}4$ model samples. For PatchCore scoring, we used a top-$q{=}0.03$ percentile and $k{=}3$ nearest neighbors. The gating threshold $\tau_z$ was dynamically relaxed from 1.0 to 1.5 when no safe candidates were found in the uncertainty-aware selection step. All hyperparameters were empirically chosen to balance computational efficiency, convergence stability, and robustness across datasets.

\subsection{Datasets} 
We evaluate the proposed framework on three medical imaging datasets:
\begin{itemize}
\item {\em Chest X-Ray (Pneumonia) Dataset.} We used publicly available \textit{Chest\_xray (Pneumonia)~\cite{kermany2018identifying}} dataset which comprises
1{,}341 normal and 3{,}875 pneumonia Chest radiographs. For our experiments, we
constructed a stratified evaluation subset of 624 images with 234 normal and 390
pneumonia cases and the rest as a training subset.
\item {\em COVID-CXR Dataset.} The COVID Chest X-ray dataset~\cite{cohen2020covidProspective}. The training set comprises 460 COVID-positive cases~\cite{santosh2021covid} and 1{,}266 normal samples, and the test set includes 116 COVID-positive and 317 normal samples. 
\item {\em Brain MRI ND\textendash5.} The Brain MRI ND\textendash5 dataset~\cite{Safwan_etal_ND5_2024}\footnote{\url{https://ieee-dataport.org/documents/brain-mri-nd-5-dataset}}
includes a training split with 8{,}389 abnormal images (excluding duplicate files as .png) and 2{,}339 normal images, and a test split of
2{,}806 axial slices (1{,}975 tumor, 831 normal).
\end{itemize}

\begin{table}[tbp]
\centering
\caption{Comparison at the Youden-optimal threshold across datasets.}
\label{tab:merged_all_datasets}
\resizebox{\textwidth}{!}{%
\begin{tabular}{llcccccccccccc}
\toprule
Dataset & Setting & ROC AUC & PR AUC & Thr* (Youden) & ACC & Precision & Recall & F1 & TN & FP & FN & TP \\
\midrule
\multirow{3}{*}{\textit{Chest\_xray (Pneumonia)}}
 & Baseline (PatchCore) & 0.6834 & 0.7656 & 0.993364 & 0.6538 & 0.7620 & 0.6487 & 0.7008 & 155 & 79 & 137 & 253 \\
 & Baseline + Adapter warmup & 0.8820 & 0.9267 & 0.952984 & 0.7628 & 0.9060 & 0.6923 & 0.7849 & 199 & 35 & 97 & 293 \\
 & Proposed (Post-IL, Oracle-free) & \textbf{0.8968} & \textbf{0.9372} & 0.953104 & \textbf{0.8093} & 0.8796 & \textbf{0.8051} & \textbf{0.8407} & 191 & \textbf{43} & \textbf{76} & \textbf{314} \\
\midrule
\multirow{3}{*}{\textit{COVID-CXR}}
 & Baseline (PatchCore) & 0.9489 & 0.8976 & 0.989044 & 0.8868 & 0.7481 & 0.8707 & 0.8048 & 283 & 34 & 15 & 101 \\
 & Baseline + Adapter warmup & 0.9961 & 0.9898 & 0.954467 & 0.9792 & 0.9652 & 0.9569 & 0.9610 & 313 & 4 & 5 & 111 \\
 & Proposed Method & \textbf{0.9982} & \textbf{0.9951} & 0.953187 & \textbf{0.9861} & \textbf{0.9583} & \textbf{0.9914} & \textbf{0.9746} & \textbf{312} & \textbf{5} & \textbf{1} & \textbf{115} \\
\midrule
\multirow{3}{*}{\textit{Brain MRI ND-5}}
 & Baseline (PatchCore)& 0.6041 & 0.7539 & 0.961838 & 0.6216 & 0.7695 & 0.6608 & 0.7109 & 440 & 391 & 671 & 1304 \\
 & Baseline + Adapter warmup & 0.6632 & 0.8174 & 0.966118 & 0.7327 & 0.8083 & 0.8132 & 0.8107 & 450 & 381 & 369 & 1606 \\
 & Proposed Method & \textbf{0.7269} & \textbf{0.8211} & 0.970559 & \textbf{0.7445} & \textbf{0.8086} & \textbf{0.8344} & \textbf{0.8213} & \textbf{441} & \textbf{390} & \textbf{327} & \textbf{1648} \\
\bottomrule
\end{tabular}}
\label{tab:compare_dataset}
\end{table}

\begin{figure}[tbp]
    \centering
    \begin{tabular}{cc}
        \includegraphics[width=0.5\linewidth]{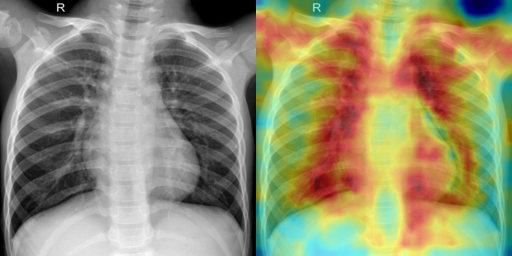} &
        \includegraphics[width=0.5\linewidth]{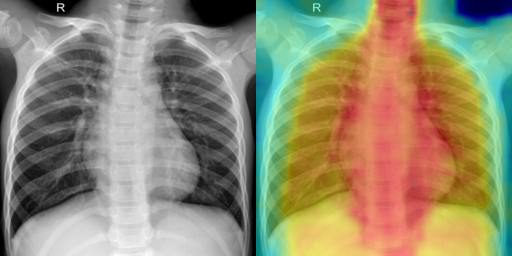}
    \end{tabular}
    \caption{Corresponding heatmaps for baseline (left) and proposed IL approach (right) on Chest\_xray dataset.}
    \label{fig:hetamap}
\end{figure}

\subsection{Results Analysis}
\begin{figure}[tbp]
    \centering
    \begin{tabular}{cc}
    \includegraphics[height=3.8cm,keepaspectratio]{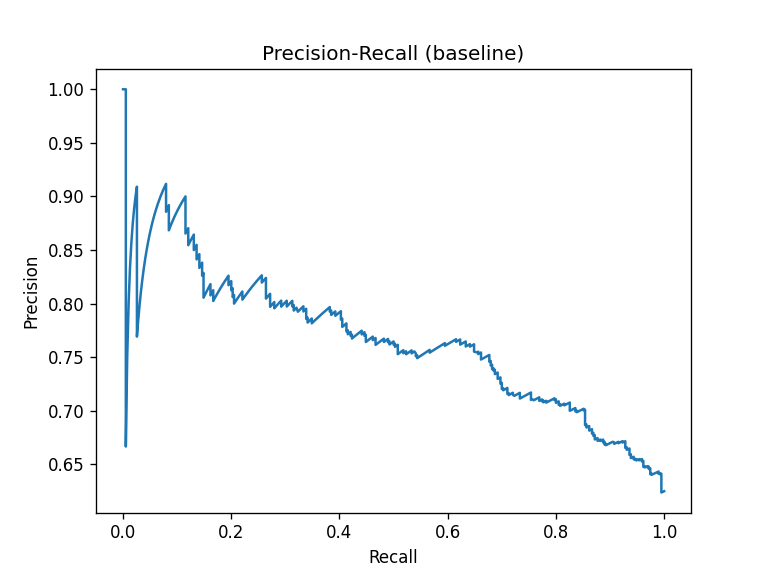} &
  \includegraphics[height=3.8cm,keepaspectratio]{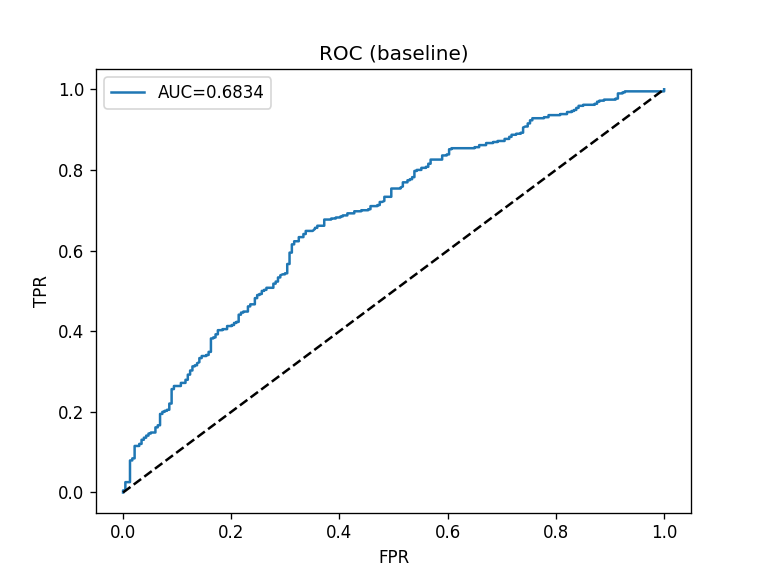}\\
  a) Baseline & b) Baseline\\
  \includegraphics[height=3.8cm,keepaspectratio]{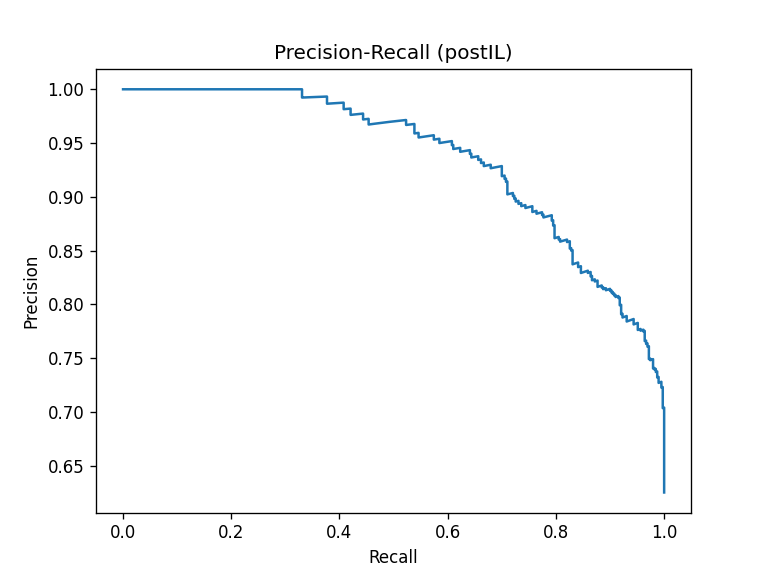} &
  \includegraphics[height=3.8cm,keepaspectratio]{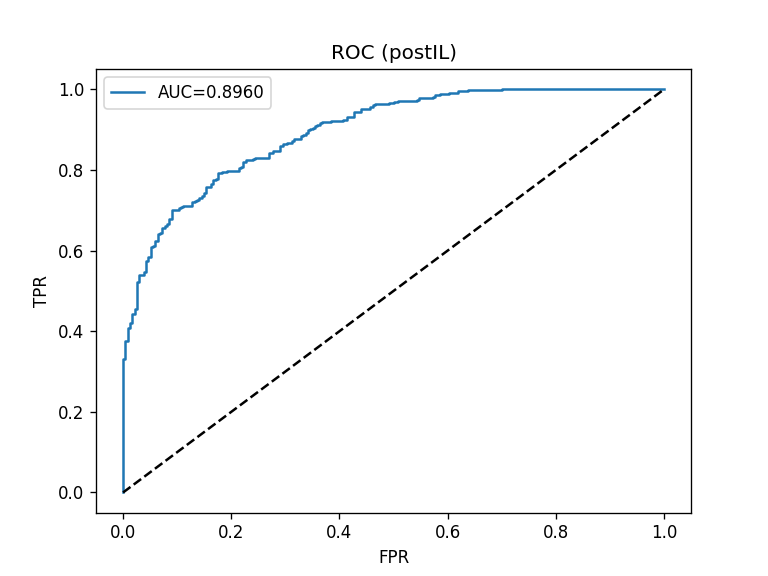}\\
  b) After IL & After IL
  \end{tabular}
    \caption{PR and ROC curves before and after IL on the Chest\_xray dataset.}
    \label{fig:pr_roc_postAL}
\end{figure}

\begin{figure}[tbp]
    \centering
    \begin{tabular}{cc}
        \includegraphics[height=4cm,keepaspectratio]{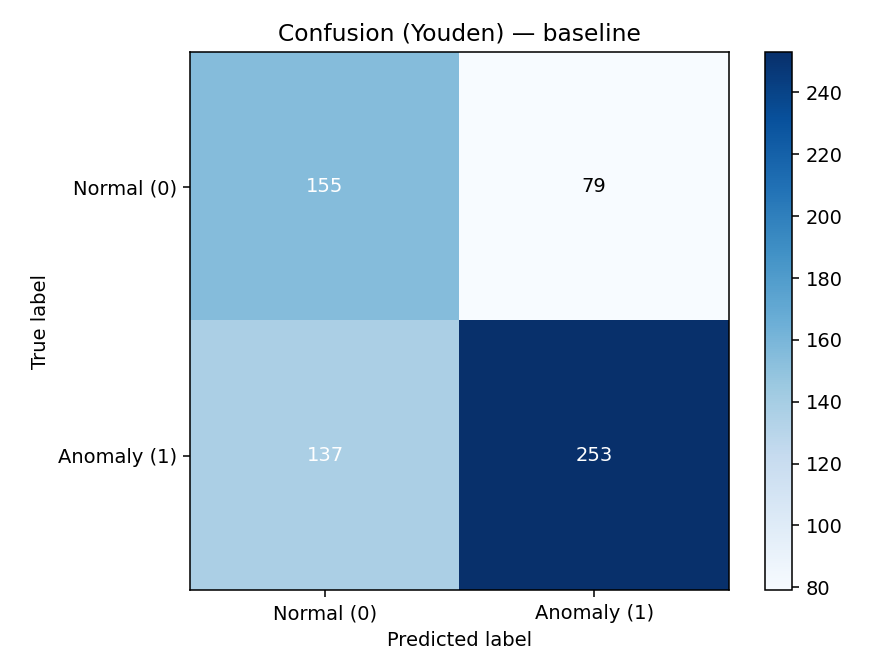} &  
        \includegraphics[height=4cm,keepaspectratio]{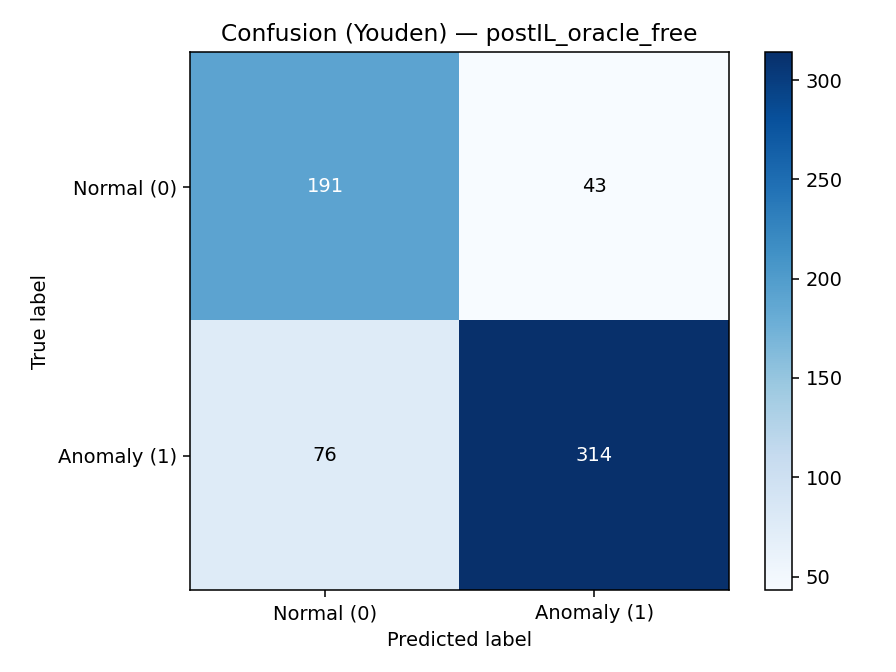}\\
         a) Baseline & b) After IL
    \end{tabular}
    \caption{Confusion matrices before and after IL. IL improves performance.}
    \label{fig:conf_matrices}
\end{figure}

\textit{a) Chest X-Ray (Pneumonia) Dataset.} We compare a baseline PatchCore system (memory from the full training set, no IL) against our proposed IL strategy using the Chest X-Ray dataset, which begins with 30\% seed normal samples. Evaluation is performed on a held-out test set with mixed labels. 
As shown in Table~\ref{tab:compare_dataset}, IL substantially improves performance across multiple metrics. ROC-AUC increases from 0.6834 (baseline) to 0.8968, while PR-AUC improves from 0.7656 to 0.9372. Accuracy also rises from 0.6538 to 0.8093. Precision improves from 0.7620 to 0.8796, with recall increasing from 0.6487 to 0.8051, resulting in an F1 gain from 0.7008 to 0.8407. In addition to these improvements, false positives are reduced from 79 to 43 while true positives increase from 253 to 314. Fig.~\ref{fig:hetamap} illustrates the heatmap of the baseline and proposed method, which shows the proposed method focusing better on the important regions. We depicted the baseline and  proposed methods’ PR and ROC curves in 
Fig.~\ref{fig:pr_roc_postAL}. For a clearer illustration of the number of correctly predicted samples compared to baseline, we present the confusion matrices in Fig.~\ref{fig:conf_matrices}.

\begin{figure}[tbp]
    \centering
    \begin{tabular}{cc}
    \includegraphics[width=0.34\linewidth,height=3.4cm]{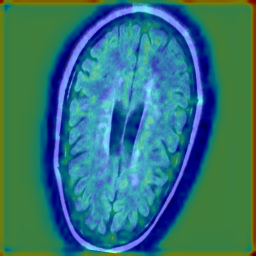} &
    \includegraphics[width=0.34\linewidth,height=3.4cm]{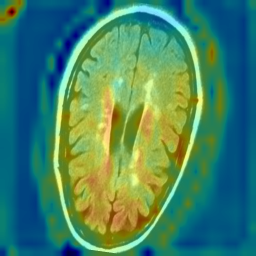}
    \end{tabular}
    \caption{Brain MRI dataset representative heatmaps before (left) and after IL (right).}
    \label{fig:heatmap-MRI}
\end{figure}

\bigskip
\noindent\textit{b) COVID\-CXR Dataset.} Table~\ref{tab:compare_dataset} reports results on the COVID Chest X-ray dataset.
Our incremental learning approach delivers substantial improvements over the baseline. ROC AUC improves from 0.9489 to 0.9982, and PR AUC rises from 0.8976 to 0.9951. At the Youden-optimal threshold, accuracy increases from 0.8868 to 0.9861. Precision improves from 0.7481 to 0.9583, recall from 0.8707 to 0.9914, and F1 score from 0.8048 to 0.9746. In terms of confusion matrix entries, true negatives increase (283 $\rightarrow$ 312), false positives decrease (34 $\rightarrow$ 5), false negatives are reduced (15 $\rightarrow$ 1), and true positives rise (101 $\rightarrow$ 115). IL not only improves ROC-AUC and PR-AUC but also reduces false positives 
(from 34 to 5) while increasing true positives (from 101 to 115).

\bigskip
\noindent\textit{c) Brain MRI Dataset (ND-5):}
We evaluated our approach on the Brain MRI ND-5, compared with the baseline we used 500 , the proposed method improves all metrics: ROC-AUC rises from 0.6041 to \textbf{0.7269} and PR-AUC from 0.7539 to \textbf{0.8211}. At the Youden-optimal threshold (baseline $t{=}0.9618$, proposed $t{=}0.9706$), accuracy increases from 0.6216 to \textbf{0.7445}. The confusion matrix shows improvement from$(\mathrm{TN},\mathrm{FP},\mathrm{FN},\mathrm{TP})=(440,391,671,1304)$ to $(441,390,327,1648)$ with a drop of \textbf{344} false negative. However, \emph{Baseline + Adapter warm-up} attains ACC 0.7327, precision 0.8083, recall 0.8132, and F1 0.8107, with $(\mathrm{TN},\mathrm{FP},\mathrm{FN},\mathrm{TP})=(450,381,369,1606)$.
Further, we depicted heatmaps in Fig.~\ref{fig:heatmap-MRI} to compare the anomaly localization behavior before and after IL. The baseline PatchCore model generates widespread, low-intensity activations, suggesting limited discriminative capacity between healthy and abnormal tissue. After IL, the attention becomes significantly more focused on the lesion area, and higher saliency contrast, indicating more discriminative feature representations.

\subsection{Ablation study using SWAG, MC Dropout, and Ensemble}
\begin{figure}[tbp]
    \centering
    \includegraphics[width=0.6\linewidth]{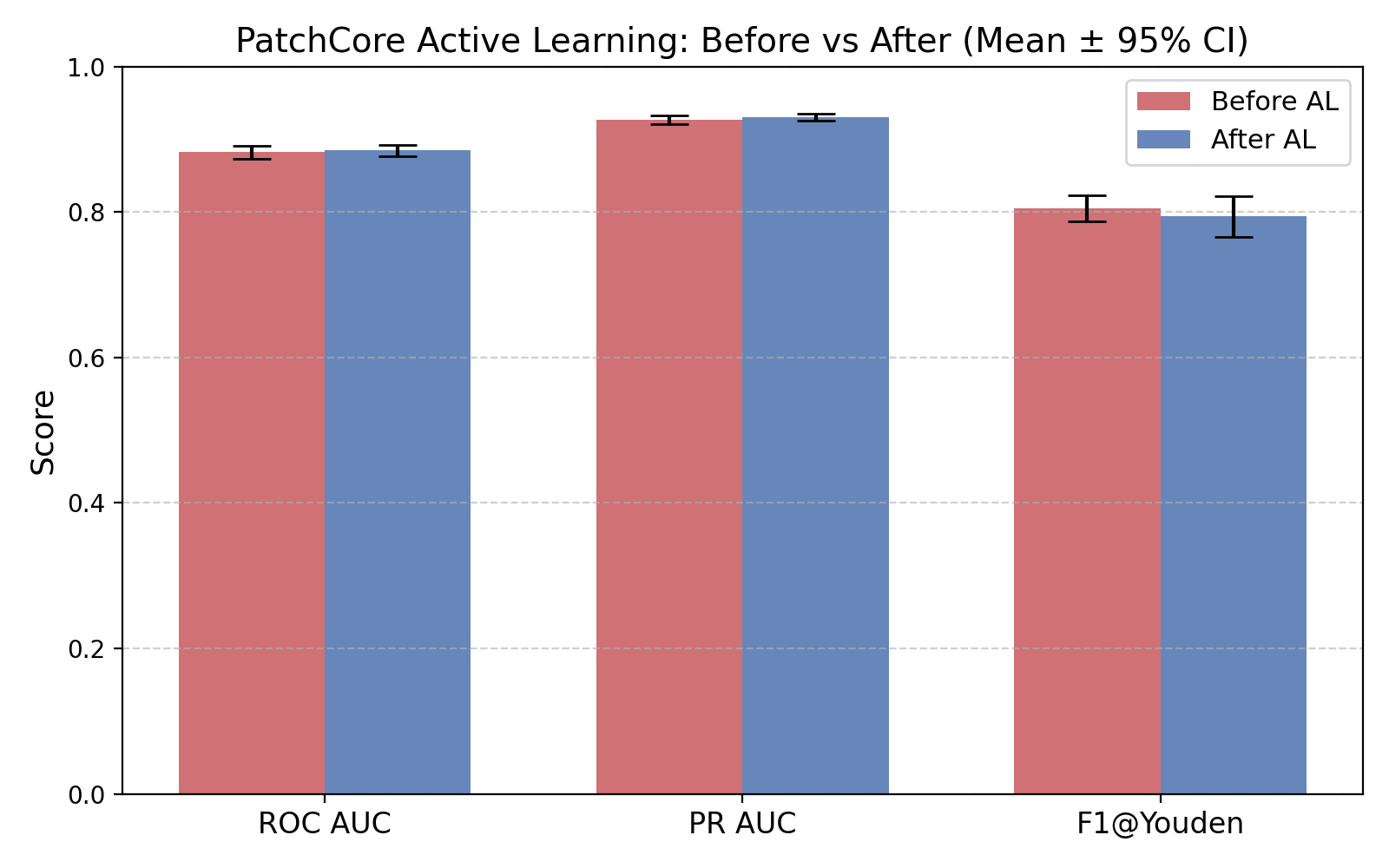}
    \caption{Comparison (Baseline+Adapter warm up) vs. after AL on Chest\_xray.}
    \label{fig:errorbar}
  \end{figure}

SWAG reaches ~0.89–0.90 ROC AUC in both oracle and oracle-free settings (Table~\ref{tab:swag_comparison}), outperforming MC Dropout and the no-uncertainty baseline. The Ensemble variant is slightly lower than SWAG. SWAG provides robust, well-calibrated uncertainty estimates; use Ensemble when computation permits for the best overall performance, and use SWAG when efficiency is a priority, particularly in incremental and unlabeled anomaly-detection scenarios.
Further, to assess the robustness of our method, we report results averaged over five independent runs with five different random seeds using the Chest\_xray dataset. The bar plots in Fig.~\ref{fig:errorbar} show the mean performance with 95\% confidence intervals for key metrics (ROC~AUC, PR~AUC, and F1@Youden). The narrow error bars indicate that our model exhibits low variance across runs, confirming the stability and reproducibility of the proposed approach.

\begin{table}[tbp]
\centering
\caption{Comparison of uncertainty estimation strategies across oracle and oracle-free settings on \texttt{Chest\_xray}.}
\begin{tabular}{lcccccc}
\hline
\textbf{Setting} & \textbf{Uncertainty} & \textbf{ROC AUC} & \textbf{PR AUC} & \textbf{ACC} & \textbf{Precision} & \textbf{F1} \\
\hline
Oracle-Free & None        & 0.8685 & 0.8934 & 0.7841 & 0.8705 & 0.7692 \\
\rowcolor{gray!10}Oracle-Free & SWAG        & 0.8968 & 0.9372 & 0.8093 & 0.8796 & 0.8407 \\
Oracle-Free & MC Dropout  & 0.8438 & 0.8797 & 0.7710 & 0.8601 & 0.7692 \\

Oracle-Free & Ensemble & 0.8828 & 0.9021 & 0.8060 & 0.8800 & 0.8380 \\
\hline
Oracle-Based & None        & 0.8372 & 0.8733 & 0.7683 & 0.8642 & 0.7561 \\
\rowcolor{gray!10}Oracle-Based & SWAG        & 0.8960 & 0.9366 & 0.8045 & 0.8829 & 0.8351 \\
Oracle-Based & MC Dropout  & 0.8099 & 0.8608 & 0.7604 & 0.8534 & 0.7521 \\

Oracle-Based & Ensemble & 0.8594 & 0.8893& 0.7950 & 0.8790 & 0.8320 \\
\hline
\end{tabular}
\label{tab:swag_comparison}
\end{table}

\subsection{Failed Case Analysis}
\begin{figure}[tbp]
    \centering
      \includegraphics[width=0.35\linewidth,height=3.8cm]{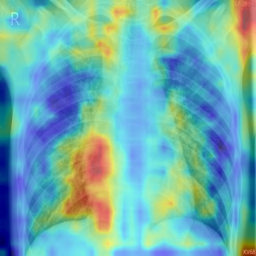}
      \includegraphics[width=0.35\linewidth,height=3.8cm]{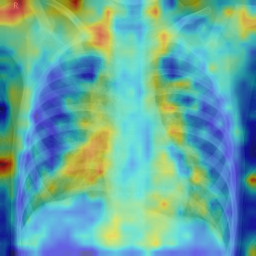}
    \caption{Failure cases on Chest\_xray, using heatmaps.}
    \label{fig:failed_case}
\end{figure}

To illustrate failure cases of an anomaly detection model for bacterial pneumonia, we illustrated the heatmaps in Fig.~\ref{fig:failed_case}. In the first example, the model misclassifies normal anatomical structures such as the cardiac border and diaphragm as abnormal regions, resulting in a false positive. In the second example, it fails to detect subtle and diffuse pulmonary opacities, leading to a false negative. These errors highlight the model’s sensitivity to normal intensity variations and its limited ability to recognize low-contrast infections.


\subsection{Limitations}
Despite strong results on standard medical-imaging benchmarks, our framework has limitations that motivate future work. Our evaluation is primarily within the medical domain; generalization to non-medical settings (e.g., industrial inspection, cybersecurity) remains untested and may stress the distance-based, uncertainty-gated admission mechanism. This is beyond the scope of this paper. 

The current two-class setup may underrepresents real-world anomaly diversity (temporal shifts, contextual changes, cross-modal inconsistencies). Long-term deployment also risks domain shift: $z$-score gates calibrated on the seed set can drift, and SWAG’s epistemic estimates may degrade across incremental rounds. Future extensions, such as multimodal inputs, drift-aware gate recalibration, and stronger uncertainty modeling may helps us to improve adaptability and generalization beyond medical imaging. Notably, many competing methods do not fit our setting: they depend on labeled replay or oracle feedback, require periodic full-dataset retraining, or assume a static independent-and-identically-distributed (IID) stream assumptions that conflict with oracle-free incremental learning. Overall, our approach offers a practical step toward label-free, uncertainty aware incremental anomaly detection.

\section{Conclusion}
In this work, we presented an unsupervised and oracle-free framework for incremental anomaly learning in medical imaging. By combining SWAG-based uncertainty estimation with distance-based screening and lightweight adapter updates, the system learns what it does not know, \textit{i.e.},  continuously refining its model of normality without supervision. By starting from a small seed of normal samples and iteratively refining the model with SWAG-guided incremental learning, our method improves detection performance without requiring any labeled anomalous data. This makes it highly practical for real-world medical imaging scenarios, where anomalies are rare, difficult to annotate, or entirely absent in training. Through extensive experiments on three publicly available medical datasets, we demonstrated that our approach consistently outperforms a baseline PatchCore system across a wide range of metrics, including ROC-AUC, PR-AUC, accuracy, precision, recall, and F1 score. Importantly, incremental learning not only improved detection sensitivity but also reduced false positives, yielding more reliable and clinically useful anomaly localization. Looking forward, our framework can be extended to other medical and industrial domains where labeled anomalies are scarce. Future work will explore integrating more advanced uncertainty quantification strategies, adapting the approach to multi-class anomaly settings, and evaluating its applicability in real-time deployment scenarios.

\section*{Acknowledgment}
This work was supported by the National Science Foundation under Grant No. \href{https://www.nsf.gov/awardsearch/showAward?AWD_ID=2346643}{\#2346643}, the U.S. Department of Defense under Award No. \href{https://dtic.dimensions.ai/details/grant/grant.14525543}{\#FA9550-23-1-0495}, and the U.S. Department of Education under Grant No. P116Z240151.
Any opinions, findings, conclusions or recommendations expressed in this material are those of the author(s) and do not necessarily reflect the views of the National Science Foundation, the U.S. Department of Defense, or the U.S. Department of Education.





\bibliographystyle{IEEEtran}
\bibliography{IEEEabrv,Bibliography}
%

\vfill


\end{document}